\pdfoutput=1

\documentclass[11pt]{article}

\usepackage[final]{paper}

\usepackage{times}
\usepackage{latexsym}

\usepackage[T1]{fontenc}

\usepackage[utf8]{inputenc}

\usepackage{microtype}

\usepackage{inconsolata}

\usepackage{graphicx}

\usepackage{fontawesome5}

\usepackage{booktabs}
\usepackage{multirow}
\usepackage{float}
\usepackage[normalem]{ulem}
\useunder{\uline}{\ul}{}

\title{Leveraging Synthetic Audio Data for \\ End-to-End Low-Resource Speech Translation}

\author{Yasmin Moslem \\
    \normalsize{Bering Lab}
  }

\begin{document}
\maketitle
\begin{abstract}
This paper describes our system submission to the International Conference on Spoken Language Translation (IWSLT 2024) for Irish-to-English speech translation. We built end-to-end systems based on Whisper, and employed a number of data augmentation techniques, such as speech back-translation and noise augmentation. We investigate the effect of using synthetic audio data and discuss several methods for enriching signal diversity. 
\end{abstract}

\section{Introduction}

Resource scarcity and the scattered nature of the data are crucial challenges for low-resource languages \citep{Lankford2021-Irish,Haddow2022-LowResourceMT,SeaCrowd2024}. In this sense, Irish is considered a low-resource language and significantly lacking in speech and language tools and resources \citep{Barry2022-Irish, Lynn2020-Irish}. Researchers have been employing various data augmentation techniques to improve the quality of low-resource textual machine translation (MT) systems. Among these techniques is using synthetic data generated by back-translation \citep{Sennrich2016-BT, Edunov2018-BT, Dowling2019-Irish, Poncelas2019-BT, Haque2020-Terminology}, or large language models \citep{Moslem2022-MT-LM}. Similarly, in the area of speech, \citet{Lee2023-SyntheticAudio} showed that models trained solely on synthetic audio datasets can generalize their performance to human voice data. Nevertheless, \citet{Guo2023-SyntheticDiversity} revealed a consistent decrease in the diversity of the outputs of language models trained on synthetic textual data. We observe that leveraging synthetic audio data generated by text-to-speech (TTS) models can be beneficial for training speech translation models, especially for low-resource languages. However, it can lack the diversity found in authentic audio signals in terms of pitch, speed, and background noise.


Speech translation systems can be cascaded systems or end-to-end systems \citep{Agarwal2023-IWSLT}. Cascaded systems use two models, one for automatic speech recognition (ASR) and one for textual machine translation (MT). End-to-end speech translation systems use one model for the whole process; hence, it is more challenging. In this work, we present end-to-end speech translation models.

In addition to describing our system submitted to IWSLT 2024, this work presents the following contributions:
\begin{itemize}
    \item Showcasing ``speech back-translation'' as an effective data augmentation technique for speech translation.~In other words, just as back-translation can improve the output quality of text-to-text MT, generating source-side \mbox{synthetic} audio data can considerably enhance the performance of speech translation systems, especially for low-resource languages.
    \item Introducing a collection of datasets for Irish-to-English speech translation, three of which comprise 196 hours of synthetic audio.
    \item Exploring diverse training settings and data processing techniques such as noise augmentation and voice audio detection (VAD).
    \item Releasing versions of Whisper models, specifically fine-tuned for Irish-to-English speech translation.
\end{itemize}

\setlength{\tabcolsep}{4.5pt} 

\begin{table*}[ht]
\centering
\begin{tabular}{lllrrr}
\toprule
\textbf{Dataset} & \textbf{Audio} & \textbf{Translation} & \textbf{Train Hours (H:M)} & \textbf{Train Segments} & \textbf{Test Segments} \\ \midrule
{\small\faUserCheck} IWSLT-2023 & Authentic & Authentic & 8:25  & 8,598  & 347 \\
{\small\faUserCheck} FLEURS & Authentic & Authentic & 16:45  & 3,991 & 0 \\
{\small\faUserCheck} Bitesize & Authentic & Authentic & 5:15 & 6,149 & 0 \\
{\small\faUserCheck} SpokenWords & Authentic & MTed & 3:02  & 10,925 & 0 \\
{\small\faGlobe}\, EUbookshop & Synthetic & Authentic & 159:45 & 67,268 & 0 \\
{\small\faComments} Tatoeba & Synthetic & Authentic & 2:39 & 3,966 & 0 \\
{\small\faWikipediaW} Wikimedia & Synthetic & Authentic & 34:23 & 15,090 & 0 \\ 
\midrule
\multicolumn{3}{l}{Authentic ({\tiny\faUserCheck}) } & 33:27 & 29,663 & 347 \\ \midrule
\multicolumn{3}{l}{Synthetic ({\tiny\faGlobe} {\tiny\faComments} {\tiny\faWikipediaW}) } & 196:47 & 86,324 & 0 \\ \midrule
\multicolumn{3}{l}{Authentic ({\tiny\faUserCheck}) + Synthetic ({\tiny\faComments} {\tiny\faWikipediaW}) } & 70:29 & 48,719 & 347 \\ \midrule
\multicolumn{3}{l}{Authentic ({\tiny\faUserCheck}) + Synthetic ({\tiny\faGlobe} {\tiny\faComments} {\tiny\faWikipediaW}) } & 229:14 & 115,987 & 347 \\ \bottomrule
\end{tabular}
\caption{Data Statistics: ``Audio'' and ``Translation'' columns refer to whether the data is human-generated or machine-generated. ``Train Hours'' and ``Train Segments'' refer to the size of the training data in terms of duration and number of utterances, respectively. Finally, ``Test Segments'' refer to the number of utterances in the test dataset.}
\label{tab:data}
\end{table*}

\section{Authentic Data}
\label{sec:authentic}

The organizers of the IWSLT shared task, provided the IWSLT-2023 dataset, which consists of training, dev, and test portions. We used both the training and dev portions for training, and the test portion for evaluation. We also used the Irish portion of the FLEURS datasets. Moreover, we employed the bilingual audio-text data available at the Bitesize website for teaching Irish.\footnote{\url{https://huggingface.co/datasets/ymoslem/BitesizeIrish-GA-EN}}

\section{Synthetic Data}
\label{sec:synthetic}

This section explains diverse approaches for creating synthetic data for speech translation. We describe each approach, as well as its advantages and disadvantages.

\subsection{Machine Translation}

When both audio and transcription are available, but there is no translation, forward MT can be useful as a data augmentation technique. However, there is the risk of feeding incorrect target translations into the training process.  Forward MT is more sensitive to the quality of the system used to produce the synthetic data. Compared to back-translation, biases and errors in synthetic data are intuitively more problematic in forward-translation, since they directly affect the gold labels \citep{Bogoychev2019-Translationese}. Hence, the used MT system must be of high quality.

We automatically translated the Irish portion of the Spoken Words dataset  into English using the Google Translation API. For quality considerations, we decided to use this dataset for training only, but not for evaluation.
The dataset consists of 10,925 utterances. Some words are spoken by multiple narrators.\footnote{\url{https://huggingface.co/datasets/ymoslem/SpokenWords-GA-EN-MTed}}

\begin{figure*}[thp]
\centering
  \includegraphics[width=0.7\linewidth]{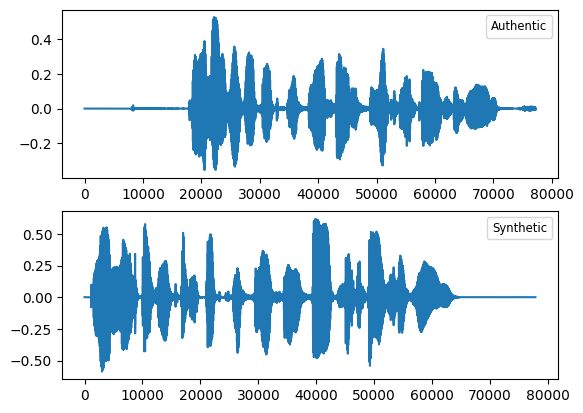}
  \caption{Comparing authentic (top) and synthetic (bottom) audio signals}
  \label{fig:authentic-synthetic}
\end{figure*}

\subsection{Synthetic Audio Data}
\label{sec:speech-back-translation}

OPUS \citep{Tiedemann2012-OPUS} hosts several bilingual textual datasets. We extracted portions of the Tatoeba, Wikimedia, and EUbookshop datasets, comprising 1,983, 7,545 and 33,634 segments, respectively.~We extensively filter the datasets based on the following criteria: removing duplicates, removing segments longer than 30 words,\footnote{\url{https://github.com/ymoslem/MT-Preparation}} language detection with fastText \citep{Joulin2017-fastText} (both sides),  and Seamless toxicity filtering \citep{Barrault2023-SeamlessM4T}. Finally, we used Azure Speech service to generate two sets of audio data, one with a female voice (OrlaNeural) and the other with a male voice (ColmNeural). As an outcome of this process, we introduce three new datasets, Tatoeba-Speech-Irish,\footnote{\url{https://huggingface.co/datasets/ymoslem/Tatoeba-Speech-Irish}} Wikimedia-Speech-Irish,\footnote{\url{https://huggingface.co/datasets/ymoslem/Wikimedia-Speech-Irish}} and EUbookshop-Speech-Irish,\footnote{\url{https://huggingface.co/datasets/ymoslem/EUbookshop-Speech-Irish}} which together comprise 196 hours of synthetic audio. Table \ref{tab:data} illustrates the statistics of our datasets.

\subsection{Audio Signal Processing Augmentation}
\label{sec:noise-vad-augmentation}

Synthetic audio data generated by TTS models can have different characteristics than authentic audio. In addition to quality considerations, we observe that among the features that distinguish data generated by TTS systems from authentic data are: 1)~lack of noise, and 2) silence differences.

\paragraph{Lack of noise:} TTS systems try to mimic studio settings, and produce very clean audio signals. However, authentic audio signals can include all sorts of environmental noise, ranging from white noise to background voices of people and cars. Even in studio settings, some breath signals can occur unless the audio is extensively edited.

\paragraph{Silence variances:} All the synthetic audio signals we generated start at a similar point, with almost no silence at the beginning of the audio (probably to facilitate mixing tracks). However, authentic audio signals can start at any point depending on the recording and processing settings, or whether a signal is truncated from a longer one.

Figure \ref{fig:authentic-synthetic} illustrates an example sentence from the Common Voice dataset uttered by a human female (non-studio settings) and its synthetic equivalent generated by Azure TTS system.\footnote{Voice name: “\textit{Microsoft Server Speech Text to Speech Voice (ga-IE, OrlaNeural)}”} The Irish sentence represented here is “\textit{Go raibh maith agaibh as ucht na fíorchaoin fáilte a d’fhear sibh romham.}” It can be translated into English as “\textit{Thank you all for that very generous welcome.}” The authentic signal has more noise (both white background noise and sounds of starting/stopping the recording software), while the synthetic signal does not show any noise occurrence. Moreover, unlike the authentic signal, the synthetic data starts almost immediately. Another observation is that this specific authentic signal has a lower volume than synthetic signals.

\subsubsection{Voice Activity Detection}
\label{sec:vad}

One of the most common audio preprocessing techniques is Voice Audio Detection (VAD). The main idea of VAD is to remove low-amplitude samples from an audio signal. Low-amplitude samples might represent science or noise samples of audio signals, which usually occur at the beginning and end of  an audio signal, but can also happen in the midst of longer audio signals. In its basic form, this can be achieved by removing any sample below an absolute value of a threshold (e.g. \textpm~0.001). However, advanced models like \textit{Silero VAD}\footnote{\url{https://github.com/snakers4/silero-vad}} can be used  as part of the \textit{torchaudio} framework, and include more sophisticated options (e.g. minimum silence duration) to avoid removing important low-amplitude samples like breath and natural silent durations.

During training, data processed with VAD can either substitute the original data or augment it, i.e. both processed and unprocessed data can be used during training. In one of our experiments (cf. Section \ref{sec:experimentss}), we used basic VAD with a threshold of \textpm~0.001 as a data augmentation technique. When basic VAD is used (i.e. without taking a minimum silence duration into account), this can also speed up the audio signal; in other words, the utterance is spoken faster. At inference time of all the models, we used Silero VAD within Faster-Whisper based on CTranslate2 \citep{Klein2020-OpenNMT}.

\subsubsection{Noise Augmentation}
\label{sec:noise}

Mimicking the effect of white noise can take diverse forms, ranging from using real noise to generating random arrays. To simulate light white noise, we generated a random array with a distribution scale 0.002 and added it to all the audio signals in the dataset.


\renewcommand{\arraystretch}{1.4} 

\begin{table*}[ht]
\centering
\begin{small}
\begin{tabular}{@{}cllrccccc@{}}
\toprule
\textbf{Whisper}        & \textbf{Model} & \textbf{Datasets}  & \textbf{Data Size} & \textbf{BLEU ↑}  & \textbf{chrF++ ↑} & \textbf{WER ↓}   & \textbf{Semantic 1 ↑} & \textbf{Semantic 2 ↑} \\ \midrule
\multirow{4}{*}{Medium} & A              & authentic          & 29,663             & 32.38          & 48.95           & 58.85          & 62.09               & 63.28               \\
                        & B              & A + synthetic (2d) & 48,719             & {\ul 36.34}    & {\ul 54.08}     & {\ul 53.35}    & {\ul 68.31}         & {\ul 69.93}         \\
                        & \textbf{B++}            & A + synthetic (3d) & 115,987            & \textbf{38.41} & \textbf{57.18}  & \textbf{51.10} & \textbf{69.72}      & \textbf{71.13}      \\
                        & C              & B + augmented      & 146,157            & 34.09          & 51.40           & 55.83          & 64.26               & 65.56               \\ \bottomrule
\end{tabular}
\end{small}
\caption{Evaluation Results: Model B++ that uses both authentic data and 3 synthetic audio datasets achieved the best results across all the systems. The results show that augmenting the training data with synthetic audio (i.e. Model B and Model B++) outperforms using authentic data only (Model A), while further signal processing augmentation with white noise and VAD (Model C) did not help. Moreover, increasing the amount of high-quality synthetic audio data in Model B++ resulted in better quality than Model B that uses a less amount of synthetic data.}
\label{tab:results}
\end{table*}

\section{Experiments}
\label{sec:experimentss}

Our experiments fine-tune Whisper \citep{Radford2022-Whisper} for the task of Irish-to-English speech translation. We experiment with a number of data augmentation techniques, such as speech back-translation (source-side synthetic audio data generation), and audio data augmentation with noise and VAD.

\subsection{Speech Back-Translation}

By the term ``speech back-translation'', we refer to generating source-side synthetic audio for data augmentation of speech translation systems, in the same manner that back-translation is employed in text-to-text MT systems. Section  \ref{sec:speech-back-translation} explains how we created these synthetic audio datasets. In this set of experiments, we built 3 systems by fine-tuning Whisper Medium. We use different types of datasets as outlined by Table \ref{tab:data}.

\begin{itemize}
    \item \textbf{Model A}: It uses the authentic data only, namely IWSLT-2023 dataset, FLEURS, Bitesize, and SpokenWords.
    \item \textbf{Model B}: It uses the same authentic data used in Model A as well as two synthetic audio datasets, namely Tatoeba-Speech-Irish, and Wikimedia-Speech-Irish.
    \item \textbf{Model B++}: In addition to the authentic and two synthetic datasets used in the aforementioned models, Model B++ uses a third synthetic dataset, namely EUbookshop-Speech-Irish.
\end{itemize}

\subsection{Noise and VAD Augmentation}

\begin{itemize}
    \item \textbf{Model C}: It uses the same data as Model B, as well as two versions of the data augmented with basic VAD, and white noise. In other words, we fine-tuned Whisper-Medium on all the authentic data and two synthetic data as well as two augmented datasets, one with low-amplitude sample removal, and one with noise augmentation, as described in Section \ref{sec:synthetic}.
\end{itemize}

\subsection{Training Arguments}

We tried different learning rates and warm-up values. Specifically, we experimented with warm-up ratios 0\%, 1\%, and 3\% out of 3000 steps, which corresponds to 0, 30, 90 warm-up steps, respectively. As Table \ref{tab:warmup-ratio-no-vad} and Table \ref{tab:warmup-ratio-vad} demonstrate, when fine-tuning Whisper Small, changing the warm-up ratio does not seem to lead to a consistent improvement for the first two sizes of data used in Model~A and Model~B. However, increasing the warm-up ratio to 3\% when the size of data is larger as in Model~C, seems to slightly improve the performance. For the learning rate, we used 1e-4 across all the experiments for the sake of consistency. The batch size was decided based on the compute capacity of one A100-SXM4-80GB GPU. Hence, we used a batch size of 64 examples when fine-tuning Whisper Small and a batch size of 16 examples when fine-tuning Whisper Medium. The max length of generation was set to 225. As this is an Irish-to-English translation task, both the tokenizer language and model generation language were set to English. We train the main models with Whisper Medium for at least two epochs, and save the best performing checkpoint based on the chrF++ score on the validation dataset. Section \ref{sec:results} elaborates on the results of these experiments.

\subsection{Training Epochs}

As we reported in the previous section, we used 3000 steps for all the experiments with Whisper Small, as further training did not seem to improve the output quality when more than one epoch of data is already reached. However, Whisper Medium was trained with a smaller batch size due to computing constrains. We wanted to see the effect of training for at least two epochs. Hence, we report different step milestones in Table \ref{tab:epochs}. In deep learning training in general, it is a common practice to use early stopping. However, for low-resource languages, a smaller value for early stopping can result in the model not seeing the whole data, which can affect the robustness of the model. This is especially true if we are not sure if the validation dataset is well-representative of the task that the model will be actually required to tackle in the real world. While there is no one rule that applies to all cases, we recommend taking this point into consideration when training generic models for low-resource languages.


\section{Evaluation and Results}
\label{sec:results}

To evaluate our systems, we calculated BLEU \citep{Papineni2002-BLEU},  chrF++ \citep{Popovic2017-chrF++}, and TER \citep{Snover2006-TER}, as implemented in the sacreBLEU library\footnote{\url{https://github.com/mjpost/sacrebleu}} \citep{Post2018-sacreBLEU}. For semantic evaluation, we used an embedding-based approach, calculating and comparing cosine similarity between the vector embeddings of each reference and the equivalent translation generated by the model. We report the average of semantic similarity across utterances. We used two models with Sentence-Transformers \citep{Reimers2019-Sentence-BERT}, \textit{“all-mpnet-base-v2”} (Semantic~1) and \textit{“all-MiniLM-L12-v2”} \citep{Wang2020-MiniLM} (Semantic~2). As we fine-tuned all the models for approximately two epochs, we report the evaluation of the best performing checkpoint.

For inference, we used Faster-Whisper \footnote{\url{https://github.com/SYSTRAN/faster-whisper}} with the default VAD arguments. We also compared the results without VAD, and found that applying VAD at inference time is better for all the models (cf. Appendix \ref{sec:appendix-arguments}). We used 5 for ``beam size'' and 2 for ``no repeat ngram size''.

As Table \ref{tab:results} shows, after fine-tuning Whisper Medium on both the authentic and synthetic audio data (Model B), there are consistent improvements across all metrics compared to when we fine-tuned it on the authentic audio data only (Model A). Moreover, Model B++ that uses three synthetic datasets outperforms Model B that uses only two synthetic datasets. This demonstrates that augmented authentic audio data with high-quality synthetic audio data can enhance end-to-end speech translation systems, especially for low-resource languages like Irish.

Model~C uses the same training data as Model~B as well as two augmented versions, one version that applies basic VAD, removing low-amplitude samples (cf. Section \ref{sec:vad}) and another version that injects white background noise into the data (cf. Section \ref{sec:noise}). Although Model C that uses noise and VAD augmented data still outperforms Model A that uses authentic training data only, both Model B and B++ that combines authentic data with synthetic data outperform Model C.

While the choice of augmentation techniques were based on manual observation of the characteristics of the authentic data and the synthetic data, the achieved improvements encourage further investigation. In the future, we would like to conduct more experiments that employ other data augmentation techniques. Moreover, we would like to measure the effect of adding synthetic audio data compared to augmenting the authentic data only. Finally, as the main purpose of this research is to understand the best practices of using synthetic audio data (i.e. data generated by TTS models) to improve speech translation quality, we will conduct further study on mimicking authentic data characteristics to enhance the effect of data augmentation with synthetic audio data.

\bibliography{paper} \clearpage

\begin{thebibliography}{26}
\providecommand{\natexlab}[1]{#1}

\bibitem[{Agarwal et~al.(2023)Agarwal, Agrawal, Anastasopoulos, Bentivogli, Bojar, Borg, Carpuat, Cattoni, Cettolo, Chen, Chen, Choukri, Chronopoulou, Currey, Declerck, Dong, Duh, Est{\`e}ve, Federico, Gahbiche, Haddow, Hsu, Mon~Htut, Inaguma, Javorsk{\'y}, Judge, Kano, Ko, Kumar, Li, Ma, Mathur, Matusov, McNamee, P.~McCrae, Murray, Nadejde, Nakamura, Negri, Nguyen, Niehues, Niu, Kr.~Ojha, E.~Ortega, Pal, Pino, van~der Plas, Pol{\'a}k, Rippeth, Salesky, Shi, Sperber, St{\"u}ker, Sudoh, Tang, Thompson, Tran, Turchi, Waibel, Wang, Watanabe, and Zevallos}]{Agarwal2023-IWSLT}
Milind Agarwal, Sweta Agrawal, Antonios Anastasopoulos, Luisa Bentivogli, Ond{\v r}ej Bojar, Claudia Borg, Marine Carpuat, Roldano Cattoni, Mauro Cettolo, Mingda Chen, William Chen, Khalid Choukri, Alexandra Chronopoulou, Anna Currey, Thierry Declerck, Qianqian Dong, Kevin Duh, Yannick Est{\`e}ve, Marcello Federico, Souhir Gahbiche, Barry Haddow, Benjamin Hsu, Phu Mon~Htut, Hirofumi Inaguma, D{\'a}vid Javorsk{\'y}, John Judge, Yasumasa Kano, Tom Ko, Rishu Kumar, Pengwei Li, Xutai Ma, Prashant Mathur, Evgeny Matusov, Paul McNamee, John P.~McCrae, Kenton Murray, Maria Nadejde, Satoshi Nakamura, Matteo Negri, Ha~Nguyen, Jan Niehues, Xing Niu, Atul Kr.~Ojha, John E.~Ortega, Proyag Pal, Juan Pino, Lonneke van~der Plas, Peter Pol{\'a}k, Elijah Rippeth, Elizabeth Salesky, Jiatong Shi, Matthias Sperber, Sebastian St{\"u}ker, Katsuhito Sudoh, Yun Tang, Brian Thompson, Kevin Tran, Marco Turchi, Alex Waibel, Mingxuan Wang, Shinji Watanabe, and Rodolfo Zevallos. 2023.
\newblock \href {https://aclanthology.org/2023.iwslt-1.1} {{FINDINGS} {OF} {THE} {IWSLT} 2023 {EVALUATION} {CAMPAIGN}}.
\newblock In \emph{{Proceedings of the 20th International Conference on Spoken Language Translation (IWSLT 2023)}}, pages 1--61, Toronto, Canada (in-person and online). Association for Computational Linguistics.

\bibitem[{Barrault et~al.(2023)Barrault, Chung, Dale, (ai), Duquenne, Elsahar, Gong, Heffernan, Hoffman, Klaiber, Chen, Licht, Maillard, Rakotoarison, Sadagopan, Wenzek, Ramakrishnan, Mourachko, Kallet, Lee, Sun, Akula, Peloquin, Huang, Yu, Ellis, Balioglu, Wood, Wang, Ropers, Gao, (fair), Kalbassi, Ye, Gonzalez, Inaguma, Schwenk, Tufanov, Kulikov, Lam, (pm Ai), Pino, Haaheim, Kao, Hasanti, Tran, Elbayad, Costa-jussa, Ramadan, El~Hachem, {\c C}elebi, Guzm{\'a}n, Tomasello, Li, Andrews, Mavlyutov, Howes, Saleem, Wang, Jain, Popuri, Tran, Vogeti, Ma, and Yang}]{Barrault2023-SeamlessM4T}
Loic Barrault, Andy Chung, David Dale, Ning~Dong (ai), Paul-Ambroise Duquenne, Hady Elsahar, Hongyu Gong, Kevin Heffernan, John Hoffman, Christopher Klaiber, Peng-Jen Chen, Daniel Licht, Jean Maillard, Alice Rakotoarison, Kaushik~Ram Sadagopan, Guillaume Wenzek, Abinesh Ramakrishnan, Alexandre Mourachko, Amanda Kallet, Ann Lee, Anna Sun, Bapi Akula, Benjamin Peloquin, Bernie Huang, Bokai Yu, Brian Ellis, Can Balioglu, Carleigh Wood, Changhan Wang, Christophe Ropers, Cynthia Gao, Daniel~Li (fair), Elahe Kalbassi, Ethan Ye, Gabriel~Mejia Gonzalez, Hirofumi Inaguma, Holger Schwenk, Igor Tufanov, Ilia Kulikov, Janice Lam, Jeff~Wang (pm Ai), Juan Pino, Justin Haaheim, Justine Kao, Prangthip Hasanti, Kevin Tran, Maha Elbayad, Marta~R Costa-jussa, Mohamed Ramadan, Naji El~Hachem, Onur {\c C}elebi, Paco Guzm{\'a}n, Paden Tomasello, Pengwei Li, Pierre Andrews, Ruslan Mavlyutov, Russ Howes, Safiyyah Saleem, Skyler Wang, Somya Jain, Sravya Popuri, Tuan Tran, Vish Vogeti, Xutai Ma, and Yilin Yang. 2023.
\newblock \href {https://ai.meta.com/resources/models-and-libraries/seamless-communication/} {{SeamlessM4T---Massively Multilingual \& Multimodal Machine Translation}}.
\newblock \emph{Meta AI}.

\bibitem[{Barry et~al.(2022)Barry, Wagner, Cassidy, Cowap, Lynn, Walsh, {\'O}~Meachair, and Foster}]{Barry2022-Irish}
James Barry, Joachim Wagner, Lauren Cassidy, Alan Cowap, Teresa Lynn, Abigail Walsh, M{\'\i}che{\'a}l~J {\'O}~Meachair, and Jennifer Foster. 2022.
\newblock \href {https://aclanthology.org/2022.lrec-1.511} {{ga{BERT} {---} an {I}rish Language Model}}.
\newblock In \emph{{Proceedings of the Thirteenth Language Resources and Evaluation Conference}}, pages 4774--4788, Marseille, France. European Language Resources Association.

\bibitem[{Bogoychev and Sennrich(2019)}]{Bogoychev2019-Translationese}
Nikolay Bogoychev and Rico Sennrich. 2019.
\newblock \href {https://arxiv.org/abs/1911.03362} {{Domain, Translationese and Noise in Synthetic Data for Neural Machine Translation}}.

\bibitem[{Dowling et~al.(2019)Dowling, Lynn, and Way}]{Dowling2019-Irish}
Meghan Dowling, Teresa Lynn, and Andy Way. 2019.
\newblock \href {https://aclanthology.org/W19-6908} {{Leveraging backtranslation to improve machine translation for {G}aelic languages}}.
\newblock In \emph{{Proceedings of the Celtic Language Technology Workshop}}, pages 58--62, Dublin, Ireland. European Association for Machine Translation.

\bibitem[{Edunov et~al.(2018)Edunov, Ott, Auli, and Grangier}]{Edunov2018-BT}
Sergey Edunov, Myle Ott, Michael Auli, and David Grangier. 2018.
\newblock \href {https://aclanthology.org/D18-1045} {{Understanding Back-Translation at Scale}}.
\newblock In \emph{{Proceedings of the 2018 Conference on Empirical Methods in Natural Language Processing}}, pages 489--500, Brussels, Belgium. Association for Computational Linguistics.

\bibitem[{Guo et~al.(2023)Guo, Shang, Vazirgiannis, and Clavel}]{Guo2023-SyntheticDiversity}
Yanzhu Guo, Guokan Shang, Michalis Vazirgiannis, and Chlo{\'e} Clavel. 2023.
\newblock \href {http://arxiv.org/abs/2311.09807} {{The Curious Decline of Linguistic Diversity: Training Language Models on Synthetic Text}}.
\newblock In \emph{{Proceedings or the Annual Conference of the North American Chapter of the Association for Computational Linguistics}}, Mexico City, Mexico.

\bibitem[{Haddow et~al.(2022)Haddow, Bawden, Barone, Helcl, and Birch}]{Haddow2022-LowResourceMT}
Barry Haddow, Rachel Bawden, Antonio Valerio~Miceli Barone, Jind{\v r}ich Helcl, and Alexandra Birch. 2022.
\newblock \href {https://aclanthology.org/2022.cl-3.6/} {{Survey of Low-Resource Machine Translation}}.
\newblock \emph{Computational Linguistics}, 06:1--67.

\bibitem[{Haque et~al.(2020)Haque, Moslem, and Way}]{Haque2020-Terminology}
Rejwanul Haque, Yasmin Moslem, and Andy Way. 2020.
\newblock \href {https://aclanthology.org/2020.icon-adapmt.4} {{Terminology-Aware Sentence Mining for {NMT} Domain Adaptation: {ADAPT}{'}s Submission to the Adap-{MT} 2020 {E}nglish-to-{H}indi {AI} Translation Shared Task}}.
\newblock In \emph{{Proceedings of the 17th International Conference on Natural Language Processing (ICON): Adap-MT 2020 Shared Task}}, pages 17--23, Patna, India. NLP Association of India (NLPAI).

\bibitem[{Joulin et~al.(2017)Joulin, Grave, Bojanowski, and Mikolov}]{Joulin2017-fastText}
Armand Joulin, Edouard Grave, Piotr Bojanowski, and Tomas Mikolov. 2017.
\newblock \href {https://aclanthology.org/E17-2068} {{Bag of Tricks for Efficient Text Classification}}.
\newblock In \emph{{Proceedings of the 15th Conference of the {E}uropean Chapter of the Association for Computational Linguistics: Volume 2, Short Papers}}, pages 427--431, Valencia, Spain. Association for Computational Linguistics.

\bibitem[{Klein et~al.(2020)Klein, Hernandez, Nguyen, and Senellart}]{Klein2020-OpenNMT}
Guillaume Klein, Fran{\c c}ois Hernandez, Vincent Nguyen, and Jean Senellart. 2020.
\newblock \href {https://aclanthology.org/2020.amta-research.9} {{The {O}pen{NMT} Neural Machine Translation Toolkit: 2020 Edition}}.
\newblock In \emph{{Proceedings of the 14th Conference of the Association for Machine Translation in the Americas (Volume 1: Research Track)}}, pages 102--109, Virtual. Association for Machine Translation in the Americas.

\bibitem[{Lankford et~al.(2021)Lankford, Alfi, and Way}]{Lankford2021-Irish}
Seamus Lankford, Haithem Alfi, and Andy Way. 2021.
\newblock \href {https://aclanthology.org/2021.mtsummit-research.5} {{Transformers for Low-Resource Languages: Is F{\'e}idir Linn!}}
\newblock In \emph{{Proceedings of Machine Translation Summit XVIII: Research Track}}, pages 48--60, Virtual. Association for Machine Translation in the Americas.

\bibitem[{Lee et~al.(2023)Lee, Jeon, Lee, Kim, and Lee}]{Lee2023-SyntheticAudio}
Jihyun Lee, Yejin Jeon, Wonjun Lee, Yunsu Kim, and Gary~Geunbae Lee. 2023.
\newblock \href {http://arxiv.org/abs/2312.01842} {{Exploring the Viability of Synthetic Audio Data for Audio-Based Dialogue State Tracking}}.
\newblock In \emph{{Proceedings of IEEE Workshop on Automatic Speech Recognition and Understanding}}, Taipei, Taiwan.

\bibitem[{Lovenia et~al.(2024)Lovenia, Mahendra, Akbar, Miranda, Santoso, Aco, Fadhilah, Mansurov, Imperial, Kampman, Moniz, Habibi, Hudi, Montalan, Ignatius, Lopo, Nixon, Karlsson, Jaya, Diandaru, Gao, Amadeus, Wang, Cruz, Whitehouse, Parmonangan, Khelli, Zhang, Susanto, Ryanda, Hermawan, Velasco, Al~Kautsar, Hendria, Moslem, Flynn, Adilazuarda, Li, Lee, Damanhuri, Sun, Qorib, Djanibekov, Leong, Do, Muennighoff, Pansuwan, Putra, Xu, Tai, Purwarianti, Ruder, Tjhi, Limkonchotiwat, Aji, Keh, Winata, Zhang, Koto, Yong, and Cahyawijaya}]{SeaCrowd2024}
Holy Lovenia, Rahmad Mahendra, Salsabil~Maulana Akbar, Lester James~V Miranda, Jennifer Santoso, Elyanah Aco, Akhdan Fadhilah, Jonibek Mansurov, Joseph~Marvin Imperial, Onno~P Kampman, Joel Ruben~Antony Moniz, Muhammad Ravi~Shulthan Habibi, Frederikus Hudi, Railey Montalan, Ryan Ignatius, Joanito~Agili Lopo, William Nixon, B{\"o}rje~F Karlsson, James Jaya, Ryandito Diandaru, Yuze Gao, Patrick Amadeus, Bin Wang, Jan Christian~Blaise Cruz, Chenxi Whitehouse, Ivan~Halim Parmonangan, Maria Khelli, Wenyu Zhang, Lucky Susanto, Reynard~Adha Ryanda, Sonny~Lazuardi Hermawan, Dan~John Velasco, Muhammad~Dehan Al~Kautsar, Willy~Fitra Hendria, Yasmin Moslem, Noah Flynn, Muhammad~Farid Adilazuarda, Haochen Li, Johanes Lee, R~Damanhuri, Shuo Sun, Muhammad~Reza Qorib, Amirbek Djanibekov, Wei~Qi Leong, Quyet~V Do, Niklas Muennighoff, Tanrada Pansuwan, Ilham~Firdausi Putra, Yan Xu, Ngee~Chia Tai, Ayu Purwarianti, Sebastian Ruder, William Tjhi, Peerat Limkonchotiwat, Alham~Fikri Aji, Sedrick Keh, Genta~Indra Winata, Ruochen
  Zhang, Fajri Koto, Zheng-Xin Yong, and Samuel Cahyawijaya. 2024.
\newblock \href {https://arxiv.org/abs/2406.10118} {{SEACrowd: A Multilingual Multimodal Data Hub and Benchmark Suite for Southeast Asian Languages}}.

\bibitem[{Lynn(2022)}]{Lynn2020-Irish}
Teresa Lynn. 2022.
\newblock \href {https://european-language-equality.eu/wp-content/uploads/2022/03/ELE___Deliverable_D1_20__Language_Report_Irish_.pdf} {{Report on the Irish Language}}.
\newblock Technical report, European Language Equality.

\bibitem[{Moslem et~al.(2022)Moslem, Haque, Kelleher, and Way}]{Moslem2022-MT-LM}
Yasmin Moslem, Rejwanul Haque, John Kelleher, and Andy Way. 2022.
\newblock \href {https://aclanthology.org/2022.amta-research.2} {{Domain-Specific Text Generation for Machine Translation}}.
\newblock In \emph{{Proceedings of the 15th biennial conference of the Association for Machine Translation in the Americas (Volume 1: Research Track)}}, pages 14--30, Orlando, USA. Association for Machine Translation in the Americas.

\bibitem[{Papineni et~al.(2002)Papineni, Roukos, Ward, and Zhu}]{Papineni2002-BLEU}
Kishore Papineni, Salim Roukos, Todd Ward, and Wei-Jing Zhu. 2002.
\newblock \href {https://aclanthology.org/P02-1040} {{{B}leu: a Method for Automatic Evaluation of Machine Translation}}.
\newblock In \emph{{Proceedings of the 40th Annual Meeting of the Association for Computational Linguistics}}, pages 311--318, Philadelphia, Pennsylvania, USA. Association for Computational Linguistics.

\bibitem[{Poncelas et~al.(2019)Poncelas, Wenniger, and Way}]{Poncelas2019-BT}
Alberto Poncelas, Gideon Maillette de~Buy Wenniger, and Andy Way. 2019.
\newblock \href {http://dx.doi.org/10.1007/978-3-031-24337-0_40} {{Adaptation of Machine Translation Models with Back-Translated Data Using Transductive Data Selection Methods}}.
\newblock In \emph{{Proceedings of the 20th International Conference on Computational Linguistics and Intelligent Text Processing CICLing 2019: Computational Linguistics and Intelligent Text Processing}}, pages 567--579, La Rochelle, France. Springer Nature Switzerland.

\bibitem[{Popovi{\'c}(2017)}]{Popovic2017-chrF++}
Maja Popovi{\'c}. 2017.
\newblock \href {https://aclanthology.org/W17-4770} {{chr{F}++: words helping character n-grams}}.
\newblock In \emph{{Proceedings of the Second Conference on Machine Translation}}, pages 612--618, Copenhagen, Denmark. Association for Computational Linguistics.

\bibitem[{Post(2018)}]{Post2018-sacreBLEU}
Matt Post. 2018.
\newblock \href {https://aclanthology.org/W18-6319} {{A Call for Clarity in Reporting {BLEU} Scores}}.
\newblock In \emph{{Proceedings of the Third Conference on Machine Translation: Research Papers}}, pages 186--191, Brussels, Belgium. Association for Computational Linguistics.

\bibitem[{Radford et~al.(2022)Radford, Kim, Xu, Brockman, McLeavey, and Sutskever}]{Radford2022-Whisper}
Alec Radford, Jong~Wook Kim, Tao Xu, Greg Brockman, Christine McLeavey, and Ilya Sutskever. 2022.
\newblock \href {https://arxiv.org/abs/2212.04356} {{Robust Speech Recognition via Large-Scale Weak Supervision}}.

\bibitem[{Reimers and Gurevych(2019)}]{Reimers2019-Sentence-BERT}
Nils Reimers and Iryna Gurevych. 2019.
\newblock \href {https://arxiv.org/abs/1908.10084} {{Sentence-BERT: Sentence Embeddings using Siamese BERT-Networks}}.

\bibitem[{Sennrich et~al.(2016)Sennrich, Haddow, and Birch}]{Sennrich2016-BT}
Rico Sennrich, Barry Haddow, and Alexandra Birch. 2016.
\newblock \href {https://aclanthology.org/P16-1009} {{Improving Neural Machine Translation Models with Monolingual Data}}.
\newblock In \emph{{Proceedings of the 54th Annual Meeting of the Association for Computational Linguistics (Volume 1: Long Papers)}}, pages 86--96, Berlin, Germany. Association for Computational Linguistics.

\bibitem[{Snover et~al.(2006)Snover, Dorr, Schwartz, Micciulla, and Makhoul}]{Snover2006-TER}
Matthew Snover, Bonnie Dorr, Rich Schwartz, Linnea Micciulla, and John Makhoul. 2006.
\newblock \href {https://aclanthology.org/2006.amta-papers.25} {{A Study of Translation Edit Rate with Targeted Human Annotation}}.
\newblock In \emph{{Proceedings of the 7th Conference of the Association for Machine Translation in the Americas: Technical Papers}}, pages 223--231, Cambridge, Massachusetts, USA. Association for Machine Translation in the Americas.

\bibitem[{Tiedemann(2012)}]{Tiedemann2012-OPUS}
J{\"o}rg Tiedemann. 2012.
\newblock \href {http://www.lrec-conf.org/proceedings/lrec2012/pdf/463_Paper.pdf} {{Parallel Data, Tools and Interfaces in {OPUS}}}.
\newblock In \emph{{Proceedings of the Eighth International Conference on Language Resources and Evaluation ({LREC}'12)}}, pages 2214--2218, Istanbul, Turkey. European Language Resources Association (ELRA).

\bibitem[{Wang et~al.(2020)Wang, Wei, Dong, Bao, Yang, and Zhou}]{Wang2020-MiniLM}
Wenhui Wang, Furu Wei, Li~Dong, Hangbo Bao, Nan Yang, and Ming Zhou. 2020.
\newblock \href {https://dl.acm.org/doi/abs/10.5555/3495724.3496209} {{MINILM: deep self-attention distillation for task-agnostic compression of pre-trained transformers}}.
\newblock In \emph{{Proceedings of the 34th International Conference on Neural Information Processing Systems}}, number Article 485 in NIPS'20, pages 5776--5788, Red Hook, NY, USA. Curran Associates Inc.

\end{thebibliography}
\appendix
\onecolumn

\section{Appendix: Arguments}
\label{sec:appendix-arguments}

\subsection{Inference VAD}

\begin{table}[!hb]
\centering
\begin{small}
\begin{tabular}{@{}lccllcc@{}}
\cmidrule(r){1-3} \cmidrule(l){5-7}
\multicolumn{1}{c}{\textbf{Argument}} & \textbf{Type} & \textbf{Value} &  & \multicolumn{1}{c}{\textbf{Argument}} & \textbf{Type} & \textbf{Value} \\ \cmidrule(r){1-3} \cmidrule(l){5-7} 
threshold & float & 0.5 &  & min\_silence\_duration\_ms & int & 2000 \\
min\_speech\_duration\_ms & int & 250 &  & window\_size\_samples & int & 1024 \\
max\_speech\_duration\_s & float & float("inf") &  & speech\_pad\_ms & int & 400 \\ \cmidrule(r){1-3} \cmidrule(l){5-7} 
\end{tabular}
\end{small}
\caption{Default VAD values of \textit{Faster-Whisper}.}
\label{tab:vad-options}
\end{table}

\subsection{Training  Warm-up Ratio}

\renewcommand{\arraystretch}{1.3} 

\begin{table}[ht]
\centering
\begin{small}
\begin{tabular}{@{}cccccccccc@{}}
\toprule
\textbf{Whisper} & \textbf{Model} & \textbf{Datasets} & \textbf{Data Size} & \textbf{Warm-up} & \textbf{BLEU} & \textbf{chrF++} & \textbf{WER} & \textbf{Semantic 1} & \textbf{Semantic 2} \\ \midrule
\multirow{9}{*}{Small} & \multirow{3}{*}{A} & \multirow{3}{*}{authentic} & \multirow{3}{*}{29,663} & 0.00 & \textbf{31.49} & 45.59 & 59.66 & 58.23 & 60.35 \\
 &  &  &  & 0.01 & 30.97 & 46.19 & \textbf{59.57} & 59.69 & 61.09 \\
 &  &  &  & 0.03 & 31.43 & \textbf{46.71} & 61.14 & \textbf{60.48} & \textbf{61.59} \\ \cmidrule(l){2-10} 
 & \multirow{3}{*}{B} & \multirow{3}{*}{A + synthetic} & \multirow{3}{*}{48,719} & 0.00 & 34.09 & \textbf{50.79} & \textbf{55.47} & \textbf{65.64} & \textbf{66.66} \\
 &  &  &  & 0.01 & 31.92 & 47.32 & 58.31 & 62.56 & 63.57 \\
 &  &  &  & 0.03 & \textbf{34.15} & 49.81 & 56.87 & 65.09 & 66.43 \\ \cmidrule(l){2-10} 
 & \multirow{3}{*}{C} & \multirow{3}{*}{B + augmented} & \multirow{3}{*}{146,157} & 0.00 & 30.75 & 45.87 & 61.37 & 60.51 & 61.98 \\
 &  &  &  & 0.01 & 32.82 & 48.31 & 57.95 & 63.26 & 64.72 \\
 &  &  &  & 0.03 & \textbf{35.07} & \textbf{50.23} & \textbf{56.73} & \textbf{63.33} & \textbf{64.80} \\ \bottomrule
\end{tabular}
\end{small}
\caption{\textbf{Comparing diverse values of warm-up ratio at training time.} Ratios are out of 3000 steps. Hence, 0.01 and 0.03 correspond to 30 steps and 90 steps, respectively. The results here are \textbf{with VAD at inference time}, using the default VAD arguments of \textit{Faster-Whisper}. The highest score in each group is displayed in a bold font.}
\label{tab:warmup-ratio-vad}
\end{table}


\setlength{\tabcolsep}{4.4pt} 

\begin{table}[hb]
\centering
\begin{small}
\begin{tabular}{@{}cccccccccc@{}}
\toprule
\textbf{Whisper} & \textbf{Model} & \textbf{Datasets} & \textbf{Data Size} & \textbf{Warm-up} & \textbf{BLEU ↑} & \textbf{chrF++ ↑} & \textbf{WER ↓} & \textbf{Semantic 1 ↑} & \textbf{Semantic 2 ↑} \\ \midrule
\multirow{9}{*}{Small} & \multirow{3}{*}{A} & \multirow{3}{*}{authentic} & \multirow{3}{*}{29,663} & 0.00 & 29.14 & 43.34 & \textbf{60.51} & 56.96 & 58.14 \\
 &  &  &  & 0.01 & 30.66 & \textbf{45.41} & 62.09 & \textbf{58.69} & \textbf{59.79} \\
 &  &  &  & 0.03 & \textbf{30.68} & 45.36 & 62.09 & 57.82 & 59.29 \\ \cmidrule(l){2-10} 
 & \multirow{3}{*}{B} & \multirow{3}{*}{A + synthetic} & \multirow{3}{*}{48,719} & 0.00 & \textbf{32.05} & \textbf{48.32} & \textbf{58.44} & \textbf{62.51} & 63.72 \\
 &  &  &  & 0.01 & 31.94 & 46.81 & 59.93 & 61.57 & 62.36 \\
 &  &  &  & 0.03 & 31.61 & 47.74 & 59.16 & 62.49 & \textbf{64.09} \\ \cmidrule(l){2-10} 
 & \multirow{3}{*}{C} & \multirow{3}{*}{B + augmented} & \multirow{3}{*}{146,157} & 0.00 & 30.51 & 44.52 & 63.48 & 59.6 & 60.71 \\
 &  &  &  & 0.01 & \textbf{32.58} & 47.65 & \textbf{59.39} & \textbf{62.86} & \textbf{63.72} \\
 &  &  &  & 0.03 & 31.89 & \textbf{48.83} & 59.84 & 62.32 & 63.17 \\ \bottomrule
\end{tabular}
\end{small}
\caption{\textbf{Comparing diverse values of warm-up ratio at training time.} Ratios are out of 3000 steps. Hence, 0.01 and 0.03 correspond to 30 steps and 90 steps, respectively. The results here are \textbf{without VAD at inference time}. The highest score in each group is displayed in a bold font.}
\label{tab:warmup-ratio-no-vad}
\end{table}

\newpage
\subsection{Training Epochs}

\setlength{\tabcolsep}{3pt} 

\begin{table}[ht]
\centering
\begin{scriptsize}

\begin{tabular}{@{}cllrccccccccc@{}}
\toprule
\textbf{Whisper} & \multicolumn{1}{c}{\textbf{Model}} & \multicolumn{1}{c}{\textbf{Datasets}} & \multicolumn{1}{c}{\textbf{Data Size}} & \textbf{Warm-up} & \textbf{Steps} & \textbf{Epoch} & \textbf{Best Epoch} & \textbf{BLEU ↑} & \textbf{chrF++ ↑} & \textbf{WER ↓} & \textbf{Semantic 1 ↑} & \textbf{Semantic 2 ↑} \\ \midrule
\multirow{10}{*}{Medium} & \multirow{2}{*}{A} & \multirow{2}{*}{authentic} & \multirow{2}{*}{29,663} & 0.03 & 2,000 & 1.08 & 1.02 & 29.14 & 47.03 & 63.17 & 60.78 & 62.11 \\
 &  &  &  & cont. & 4,000 & 2.16 & 1.83 & {\ul 32.38} & {\ul 48.95} & {\ul 58.85} & {\ul 62.09} & {\ul 63.28} \\ \cmidrule(l){2-13} 
 & \multirow{2}{*}{B} & \multirow{2}{*}{A + synthetic (2d)} & \multirow{2}{*}{48,719} & 0.03 & 4,000 & 1.31 & 1.22 & 36.02 & 53.73 & {\ul 53.26} & 66.86 & 68.16 \\
 &  &  &  & cont. & 7,000 & 2.30 & 2.27 & {\ul 36.34} & {\ul 54.08} & 53.35 & {\ul 68.31} & {\ul 69.93} \\ \cmidrule(l){2-13} 
 & \multirow{3}{*}{B++} & \multirow{3}{*}{A + synthetic (3d)} & \multirow{3}{*}{115,987} & 0.03 & 4,000 & 0.55 & 0.55 & \textbf{38.41} & \textbf{57.18} & \textbf{51.10} & \textbf{69.72} & \textbf{71.13} \\
 &  &  &  & cont. & 8,000 & 1.10 & 0.55 & $\sim$ & $\sim$ & $\sim$ & $\sim$ & $\sim$ \\
 &  &  &  & cont. & 15,000 & 2.07 & 0.55 & $\sim$ & $\sim$ & $\sim$ & $\sim$ & $\sim$ \\ \cmidrule(l){2-13} 
 & \multirow{3}{*}{C} & \multirow{3}{*}{B + augmented} & \multirow{3}{*}{146,157} & 0.03 & 4,000 & 0.44 & 0.38 & 33.46 & 50.72 & 57.59 & 63.01 & 64.56 \\
 &  &  &  & cont. & 10,000 & 1.09 & 1.05 & {\ul 34.09} & {\ul 51.4} & {\ul 55.83} & {\ul 64.26} & {\ul 65.56} \\
 &  &  &  & cont. & 19,000 & 2.08 & 1.05 & $\sim$ & $\sim$ & $\sim$ & $\sim$ & $\sim$ \\ \bottomrule
\end{tabular}

\end{scriptsize}
\caption{Investigating the effect of training for 1-2 epoch(s). It seems that smaller amounts of training data can benefit from training for 2+ while larger amounts of data can benefit from training for only 1 epoch or less. The first row of each section starts the training with warm-up ratio 0.03, then the next 1 or 2 row(s) continues training for more steps without changing any training arguments. The reported scores are for the best step, based on training validation with 100-step intervals. That is why some rows are marked with the ``$\sim$" sign, as the best step was still the same as the one reported in the previous row.}
\label{tab:epochs}
\end{table}

\end{document}